\documentclass{article}

\usepackage{microtype}
\usepackage{graphicx}
\usepackage{subcaption}
\usepackage{booktabs} 
\usepackage[english]{babel}
\usepackage{tikz}

\usepackage{adjustbox}
\usepackage{rotating}
\usepackage{subcaption}
\usepackage{mdframed} 
\usetikzlibrary{shapes, arrows, positioning}
\usepackage{pgfplots}
\usepackage{listings}
\usepackage{xcolor}
\usepackage{float}
\usepackage{stfloats}
\usepackage{csquotes}
\usepackage{tabularx}
\usepackage{multirow}
\usepackage{booktabs}  
\usepackage{caption}
\usepackage{pgfplots}
\usepgfplotslibrary{groupplots}
\usepackage{fancyhdr}

\usepackage{amsmath}
\usepackage{graphicx}
\usepackage[colorlinks=true, allcolors=blue]{hyperref}
\usepackage{hyperref}

\usepackage[accepted]{icml2026}

\makeatletter
\renewcommand{\Notice@String}{}
\makeatother

\usepackage{amsmath}
\usepackage{amssymb}
\usepackage{mathtools}
\usepackage{amsthm}

\usepackage[capitalize,noabbrev]{cleveref}

\theoremstyle{plain}

\theoremstyle{definition}

\theoremstyle{remark}

\usepackage[textsize=tiny]{todonotes}

\icmltitlerunning{MAVEN: Improving Generalization in Agentic Tool Calling}

\begin{document}

\twocolumn[
  \icmltitle{MAVEN: Improving Generalization in Agentic Tool Calling}

  \begin{icmlauthorlist}
    \icmlauthor{Omkar Ghugarkar}{corethink}
    \icmlauthor{Vishvesh Bhat}{corethink}
    \icmlauthor{Muhammad Ahmed Mohsin}{stanford}
    \icmlauthor{Asad Aali}{stanford}
  \end{icmlauthorlist}

  \icmlaffiliation{corethink}{CoreThink AI, USA}
  \icmlaffiliation{stanford}{Stanford University, Stanford, CA, USA}

  \icmlcorrespondingauthor{Vishvesh Bhat}{vish@corethink.ai}

  \vskip 0.15in
  \begin{center}
  {\small
    $^{1}$CoreThink AI, USA\\
    $^{2}$Stanford University, Stanford, CA, USA
  }
  \end{center}

  \icmlkeywords{Machine Learning, ICML}

  \vskip 0.3in
]



\printAffiliationsAndNotice{}  

\begin{abstract}
Generalization across agentic tool-calling environments remains a central challenge for reliable agentic reasoning systems. Although large language models achieve strong results on individual benchmarks, their ability to compose reasoning strategies, preserve intermediate states, and coordinate tools across domains remains underexplored. We present \textbf{MAVEN} (Modular Agentic Verification and Execution Network), a lightweight symbolic reasoning scaffold for structured decomposition, adaptive tool orchestration, and intermediate verification. We evaluate MAVEN across established tool-calling benchmarks, including BFCL v3, TauBench, Tau2Bench, AceBench, and introduce \textbf{MAVEN-Bench}, a stress-test benchmark for multi-step mathematical and physical reasoning with explicit verification and adversarial task composition. MAVEN-Bench exposes a substantial gap between partial reasoning quality and end-to-end task success; in direct MAVEN-Bench runs, MAVEN improves its GPT-OSS-120b base model from 48\% to 71\% accuracy without additional training. It also remains competitive with frontier proprietary baselines while using an open-weight backbone with an estimated cost ratio of roughly 1/10, suggesting that lightweight verification-centered scaffolds can strengthen compositional reasoning and motivate more process-aware evaluation of agents in the wild.
\end{abstract}

\begin{figure*}[t!]
\centering
\includegraphics[width=0.9\linewidth]{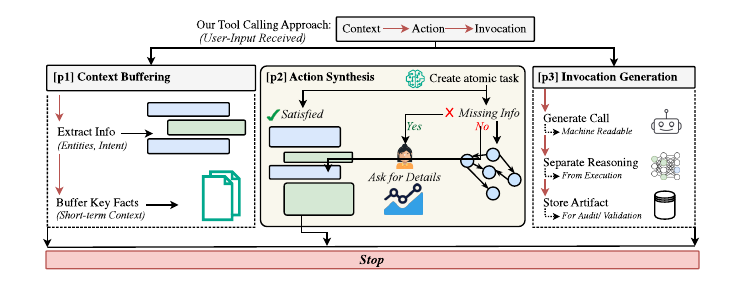}
\caption{The system processes conversational input through three stages: Context Buffering extracts and structures relevant information, Action Synthesis generates atomic, testable tasks while handling early termination and missing prerequisites, and Invocation Generation produces machine-interpretable actions with auditability, keeping reasoning and execution separated.}
\label{tc}
\end{figure*}

\section{Introduction}
Large language models (LLMs) are increasingly used as the foundation for autonomous, ``agentic'' systems that plan, reason, and interact with external tools. These systems are useful in domains where a model must decompose a user request into intermediate steps, select appropriate tools, execute those tools in the correct order, and verify partial results before producing a final answer. However, many current agentic systems remain brittle in long-horizon settings. This setting can be interpreted as a compositional reasoning problem, where agents must construct solutions by combining reusable tools and intermediate representations.

Evaluating these behaviors requires benchmarks that measure more than final-answer correctness. Existing datasets such as BFCL v3, TauBench, Tau2Bench, and AceBench evaluate important aspects of function calling, interactive tool use, and agentic task completion~\cite{bfcl_v3,tau_bench,tau2_bench,acebench}. However, strong performance on a fixed benchmark suite does not necessarily imply robust reasoning under new task structures. Models may adapt to dataset-specific formats, tool schemas, or interaction patterns, a concern related to broader benchmark robustness issues~\cite{lunardi2025robustnessreliabilitybenchmarkbasedevaluation}. This motivates evaluation settings that stress process fidelity, intermediate state management, and explicit verification.

To study these issues, we introduce \textbf{MAVEN-Bench}, a benchmark for multi-step scientific reasoning with external tools. MAVEN-Bench focuses on parameterized mathematics and physics problems that require symbolic, numeric, and verification-oriented tool calls. Each instance is designed to test whether an agent can preserve intermediate state, select appropriate tools, handle adversarial parameter regimes, and verify its own computations.

We also introduce \textbf{MAVEN} (Modular Agentic Verification and Execution Network), a scaffolded reasoning layer that encourages structured decomposition, explicit intermediate verification, and adaptive tool orchestration. We instantiate MAVEN primarily on top of GPT-OSS-120b and compare it against the GPT-OSS-120b base model.

We make three contributions. First, we introduce MAVEN-Bench, a process-aware benchmark for tool-augmented mathematical and physical reasoning. Second, we provide an evaluation protocol that records tool traces, intermediate artifacts, and verification behavior. Third, we evaluate MAVEN and show that reasoning scaffolds improve performance over the base model, particularly on tasks requiring long-horizon decomposition and intermediate verification.

\section{Related Work}

The Berkeley Function Calling Leaderboard (BFCL) v3 evaluates large language models' (LLMs) ability to invoke external functions in multi-turn, multi-step settings with explicit state tracking~\cite{bfcl_v3}. The benchmark extends prior versions by incorporating long-context reasoning and missing-function scenarios. However, its reliance on Abstract Syntax Tree (AST)-based evaluation may not fully capture semantic correctness in real-world tool usage~\cite{bfcl_critic, rabinovich2025robustnessagenticfunctioncalling, survey}. $\tau$-Bench models interactive tool usage through simulated user-agent dialogues under domain-specific constraints, evaluating both task completion and policy adherence in retail and airline~\cite{tau_bench}. While it captures structured interaction patterns, its reliance on limited domains constrains its ability to reflect real-world variability and limits evaluation of cross-domain generalization~\cite{survey, yao2022react}. $\tau^2$-Bench extends $\tau$-Bench by introducing a dual-control environment where both agent and user interact within a shared state, enabling evaluation of coordination and communication in more realistic settings~\cite{tau2_bench}. Despite improved interaction fidelity, the increased complexity introduces ambiguity in performance attribution and reduces evaluation consistency, particularly across heterogeneous task settings~\cite{survey, shinn2023reflexion}. ACEBench provides a fine-grained evaluation of function-calling behavior by categorizing tasks into Normal, Special, and Agent settings, enabling detailed analysis of parameter-level correctness and multi-step execution~\cite{acebench}. However, its dependence on LLM-based evaluation or real API execution introduces computational overhead and scalability constraints, and its predefined categories may not fully capture the diversity of real-world tool usage scenarios~\cite{survey, qin2023toolllm}.

\section{Methodology}

\begin{algorithm}[t]
\caption{MAVEN Structured Tool-Use Procedure}
\label{alg:maven}
\begin{algorithmic}[1]
\STATE \textbf{Input:} query $q$, tools $\mathcal{T}$, environment $E$
\STATE Initialize buffer $B \leftarrow \textsc{Buffer}(q)$
\WHILE{no completion signal has been produced}
    \STATE Generate subtask $a_i \leftarrow \textsc{Synthesize}(B,q)$
    \IF{$a_i$ is empty or prerequisites are missing}
        \STATE \textbf{break}
    \ENDIF
    \STATE Select tool $t_i \in \mathcal{T}$ and build invocation $u_i$
    \STATE Execute $u_i$ in $E$
    \STATE Store output, diagnostics, provenance, and verification result in $B$
    \IF{verification fails}
        \STATE Revise $a_i$ or $u_i$ using $B$
    \ENDIF
\ENDWHILE
\STATE Return final answer and audit trace
\end{algorithmic}
\end{algorithm}

\textbf{MAVEN} (Figure~\ref{tc}; Algorithm~\ref{alg:maven}) translates conversational context into structured, verifiable actions while optionally producing executable invocations, with an explicit emphasis on minimizing unsafe side effects and maintaining auditability. The approach follows a three-stage pipeline. First, in the context buffering stage, the system extracts and organizes relevant information from the input conversation into a compact, short-lived representation that preserves salient facts and any intermediate reasoning necessary for downstream processing. Second, in the action synthesis stage, the buffered representation is used to generate an atomic and testable task description aligned with the user’s objective; this stage incorporates a bounded refinement process to ensure correctness and clarity while avoiding unnecessary iterations, and it supports early termination when no further action is required. Finally, in the invocation generation stage, the system produces a machine-interpretable action that is compatible with the execution environment once all prerequisites are satisfied; by explicitly separating reasoning from execution, the system reduces the likelihood of unintended side effects while retaining a compact audit artifact to support validation, human inspection, and post-hoc analysis. This staged design ensures a balance between reliability and computational efficiency.

\begin{figure}[t!]
\centering
\includegraphics[width=1\linewidth]{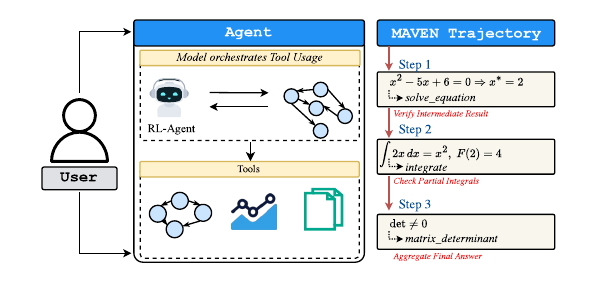}
\caption[Agent--tool orchestration for MAVEN-Bench]{Schematic of the MAVEN-Bench evaluation setup. A user supplies a multi-step math or physics problem; the Agent orchestrates calls to external tools (e.g., \texttt{solve\_equation}, \texttt{integrate}, \texttt{matrix\_determinant}, \texttt{linear\_regression}), verifies intermediate results at each step, and aggregates those results to produce the final solution. Right: an example MAVEN-Bench trajectory showing sequential, step-wise tool calls with intermediate verification and aggregation.}
\label{setup}
\end{figure}

\section{Benchmark}
\label{sec:maven}
\textbf{MAVEN-Bench} (Figure~\ref{setup}) is an evaluation ecosystem designed to measure the ability of tool-using agents to perform extended, verifiable scientific problem solving. The benchmark concentrates on three interlocking competencies: reliable orchestration of multiple specialised computational tools, disciplined preservation and inspection of intermediate state, and the explicit verification of intermediate results that together produce reproducible final outcomes. Unlike short-question corpora that can encourage rote pattern matching or single-step retrieval, MAVEN-Bench targets sustained chains of reasoning that are representative of scientific workflows where diagnostic awareness is essential. MAVEN-Bench evaluates agents on multi-step reasoning tasks under stress-test conditions where final-answer correctness alone is insufficient. It stresses symbolic, numeric, and tool-augmented reasoning across extended solution traces, measuring not only what an agent outputs but how reliably it preserves state, handles edge cases, and verifies results.

\begin{figure}[t]
\centering
\captionsetup{font=small}

\begin{mdframed}[
linewidth=0.5pt,
roundcorner=3pt,
backgroundcolor=gray!5,
innertopmargin=6pt,
innerbottommargin=6pt,
innerleftmargin=6pt,
innerrightmargin=6pt
]

\scriptsize
\textbf{Step 1: Tool Invocation and Persistence}
\begin{lstlisting}
POST /mcp/call
Body: {
  "problem_id": "MAVEN-Bench-0001",
  "step_id": "step-01",
  "tool_id": "symbolic_diff",
  "input": { "expr": "A*t^3 - B*t^2 + C*t", "wrt": "t" },
  "persist": true
}
Response: {
  "ok": true,
  "result_id": "MAVEN-Bench-0001-step-01-result",
  "output": { "expr": "3*A*t^2 - 2*B*t + C" },
  "diagnostics": { "type": "symbolic", "simplified": true }
}
\end{lstlisting}

\vspace{4pt}

\textbf{Step 2: Querying Persisted State}
\begin{lstlisting}
POST /mcp-server/mcp
Body: {
  "problem_id": "MAVEN-Bench-0001",
  "query": { "from_step": "step-01", "fields": ["output.expr"] }
}
Response: {
  "ok": true,
  "matches": [
    { "result_id": "MAVEN-Bench-0001-step-01-result",
      "output": { "expr": "3*A*t^2 - 2*B*t + C" } }
  ]
}
\end{lstlisting}

\end{mdframed}

\caption{Minimal MCP interaction example illustrating tool invocation, persistence of intermediate results, and retrieval for downstream reasoning.}
\label{mcp}
\end{figure}

\subsection{Dataset Composition and Parametric Instantiation}
MAVEN-Bench's core corpus contains one hundred canonical problem templates drawn from calculus, algebra, linear algebra, classical mechanics, thermodynamics, electromagnetism, and applied mathematics. Each template is parameterised so concrete instantiations vary in numerical regime, algebraic form, and verification requirements; small parameter changes can induce ill-conditioning, near-degenerate stationary points, or multiple algebraic branches. This design raises the bar for generalisation by requiring robust tool selection, conditioning-aware computation, and explicit verification across instantiations.

\subsection{Model Context Protocol (MCP) and Persistent State}
A defining component of MAVEN-Bench (Figure~\ref{mcp}) is the Model Context Protocol (MCP), a protocol and reference implementation that formalises how agents persist, query, and reason about intermediate results. MCP treats intermediate artifacts as first-class objects: symbolic expressions, numerics with units, solver diagnostics, and provenance metadata are each stored under explicit step identifiers and made available for later retrieval. This persistence model supports experimental questions that are otherwise difficult to study in isolation: how often should agents recompute versus reuse persisted results, which forms of intermediate representation improve downstream stability, and how does explicit provenance aid automated verification? The benchmark artifacts, MCP server, and client examples will be released upon acceptance to support reproducible evaluation while preserving the anonymity of the review process.

\subsection{Construction and Validation Pipeline}
Problems in MAVEN-Bench were produced by a multi-stage pipeline that mixes human expertise and automated validation. Domain specialists drafted seeds that require multi-concept reasoning and identified canonical tool-paths that typify safe solution strategies. These seeds were then augmented adversarially by injecting distractor terms, creating parameter regimes that stress numerical stability, and permuting algebraic forms to produce alternative but valid solution branches. Canonical traces--sequences of tool calls together with expected intermediate outputs--were executed in the MCP sandbox to generate ground-truth artifacts. Automated perturbation tests followed, running canonical paths under variations of solver tolerances, input perturbations, and ordering permutations to ensure the reference answers and verification checkpoints are stable under plausible variations. Finally, independent human reviewers audited both the canonical traces and the validation harness to ensure that the recorded traces capture meaningful, domain-relevant reasoning steps rather than artefacts.

\subsection{Annotation, Diagnostics and Failure-Mode Disclosure}
Every MAVEN-Bench instance is designed to include a rich annotation bundle. Step-level traces record the tool invoked, the exact input payload, expected outputs annotated with acceptable equivalence classes, and recommended numeric tolerances. Crucially, each persisted artifact includes diagnostic metadata such as solver status flags, convergence metrics, condition numbers, and simplification provenance. We treat failure modes as first-class information: known pitfalls (for example, divisions by small numbers, ambiguous branch cuts, or near-singular matrices) are documented alongside recommended checks. This emphasis on diagnostics and explicit failure-mode disclosure serves two purposes. Practically, it allows evaluators to distinguish between a correct-looking final answer produced by reckless numerics and a robustly verified result. Scientifically, it provides a substrate for studying how explicit diagnostic information affects agent behaviour and error-correction.

\subsection{Evaluation Protocol and Metrics}
MAVEN-Bench evaluates agents along multiple, complementary axes that reflect the multifaceted nature of reliable computation. Sub-question accuracy measures correctness at the granularity of individual verification checkpoints and intermediate deliverables. Tool selection accuracy captures whether the chosen primitive is appropriate for the subtask at hand. Trace fidelity quantifies alignment between the agent's MCP trace and the canonical reference while tolerating legitimate variations in ordering and representation. Verification score measures whether explicit checks--units consistency, second-derivative tests for extrema, or solver residual-based acceptance criteria--were both executed and correctly interpreted. Final-answer correctness employs symbolic-equivalence checks and numeric tolerances.
\subsection{On Generalization and Risk of Overfitting}
MAVEN-Bench addresses overfitting through parametric richness, multiple valid tool paths, and adversarial perturbations that change the numerical or algebraic character of instances. Because canonical traces are seldom unique, memorizing a single path performs poorly under perturbation and on metrics that reward verification and provenance. The full MCP traces, problem generators, and validation harness will be released upon acceptance.

\section{Evaluation}
\label{sec:evaluation}

We evaluate systems on MAVEN-Bench using a trace-aware harness for long-horizon, multi-step mathematics and physics problems. The harness records tool calls, intermediate artifacts, diagnostic metadata, and final answers. Evaluation emphasizes three aspects of behavior: whether the agent selects appropriate tools, whether intermediate results are correct and verified, and whether the final answer is correct under symbolic or numeric equivalence checks. We use GPT-4.1 as an automated judge for rubric-based assessment, following the broader practice of using strong LLM judges in structured evaluation settings such as HealthBench~\cite{health}. To reduce ambiguity, judge prompts use a fixed rubric and are applied to logged traces rather than free-form model outputs alone.

\subsection{Testing Protocol}

All MAVEN-Bench evaluations use a constrained execution protocol. Models are instructed to operate in a tools-only setting: computations should be performed through explicit MCP tool calls rather than implicit arithmetic or symbolic manipulation inside natural language. Each model response is limited to a single tool invocation or a termination action, which enforces sequential decomposition and allows trace-level analysis. A canonical completion signal, \texttt{PROBLEM\_COMPLETED}, is required to mark termination. The execution environment is deterministic and version-controlled so that identical tool inputs produce identical tool outputs. Violations of these constraints are handled systematically, allowing the evaluation pipeline to remain robust to imperfect adherence while preserving fairness in scoring.

\subsection{Trace Reconstruction and Rule-Violation Handling}
\label{sec:trace-recon}

Protocol violations are recorded explicitly. If a model violates the single-call constraint, omits a required MCP wrapper, or embeds a tool-like operation in natural language, the run is marked with a violation flag. Trace reconstruction is used only as a diagnostic secondary analysis: the evaluator may attempt to map the model's output into a valid sequence of MCP calls in order to understand what the model appeared to intend. Reconstructed traces are not counted as clean primary successes unless the original execution also satisfies the protocol. Primary accuracy, therefore, reflects valid executions under the stated tool-use constraints, while reconstructed accuracy, when reported, is treated as an auxiliary diagnostic.

\subsection{Judge Rubric}

The automated judge scores each completed trace using a fixed rubric with three components. The tool-use score measures whether the selected tools are appropriate, whether their inputs are well-formed, and whether the sequence of calls is sufficient for the task. The correctness score measures the mathematical or physical correctness of intermediate and final outputs under symbolic equivalence or numeric tolerance. The approach score measures whether the agent performs necessary verification checks, handles units and conditioning issues, and avoids unsupported reasoning shortcuts. The judge is given the problem statement, the full MCP trace, tool outputs, and the final answer. It is not asked to infer missing computations in the trace.

\subsection{Experimental Setup and Reproducibility}

All tools exposed through MCP are versioned and instrumented to record tool identifiers, inputs, outputs, and diagnostic metadata. Runs use fixed decoding settings, bounded time budgets, and deterministic tool execution. The benchmark records both aggregate metrics and per-instance traces to support reproducibility. Calibration of the judge rubric is performed on a held-out validation set, and human audits are used to inspect a sample of judge decisions for alignment with the written rubric. For reproducibility, all MAVEN-Bench leaderboard results are reported on 100 concrete evaluated instances per model, corresponding to one instantiated task for each of the 100 canonical MAVEN-Bench templates. The model identifiers used in the leaderboard evaluation were: \textit{gpt-5-2025-08-07} for GPT-5, \textit{o4-mini-2025-04-16} for o4-mini, \textit{o3-2025-04-16} for o3, \textit{gemini-2.5-pro-preview-05-06} for Gemini-2.5, \textit{Kimi-K2}, \textit{DeepSeek-V3.1}, and \textit{Qwen3-235B}. Per-question results for each evaluated model are included in the released results directory.

\subsection{Evaluation Outputs}

For each model, MAVEN-Bench produces per-instance scoring records with component-wise scores, completion status, protocol-violation flags, judge critiques, and trace metadata. Aggregate summaries report end-to-end accuracy, mean score, completion rate, tool-selection accuracy, verification score, and common error modes. Canonical traces and model-generated traces are retained for post-hoc analysis.

\section{Results}

\subsection{General Tool-Calling Benchmarks}

We evaluate MAVEN across several tool-calling benchmarks (Table~\ref{tab:model-scores-expanded}) to assess behavior under different interaction formats and assumptions. MAVEN is instantiated as a scaffolded reasoning layer on top of GPT-OSS-120b. We report two outcomes: direct comparisons between MAVEN and its GPT-OSS-120b base model, and reference comparisons to previously reported scores for broader context.

\begin{table}[t]
\centering
\caption{Mixed-source performance summary across tool-calling benchmarks. MAVEN results are direct runs; other model scores are reference values from prior reports~\cite{agentscaler} unless otherwise noted. Best results per row are highlighted in \textbf{bold}.}
\label{tab:model-scores-expanded}
\footnotesize
\setlength{\tabcolsep}{4pt}
\renewcommand{\arraystretch}{1.15}
\resizebox{\columnwidth}{!}{
\begin{tabular}{l l c c c c c c c c}
\toprule
\textbf{Dataset} & \textbf{Domain} & \textbf{MAVEN} & \textbf{GPT-5} & \textbf{o4-mini} & \textbf{o3} & \textbf{Kimi-K2} & \textbf{DeepSeek-V3.1} & \textbf{Qwen3-Th-235B} & \textbf{Gemini-2.5} \\
\midrule
Tau   & Airline  & \textbf{56.00} & 44.00 & 46.00 & 52.00 & 51.20 & 40.00 & 46.00 & 44.00 \\
Tau   & Retail   & 75.65 & \textbf{78.30} & 70.40 & 70.40 & 73.90 & 66.10 & 67.80 & 68.70 \\
\midrule
Tau2  & Airline  & \textbf{62.00} & \textbf{62.00} & 46.00 & 52.00 & 51.20 & 40.00 & 46.00 & 44.00 \\
Tau2  & Retail   & 77.19 & \textbf{81.10} & 70.40 & 70.40 & 73.90 & 66.10 & 67.80 & 68.70 \\
Tau2  & Telecom  & 66.67 & \textbf{96.70} & 46.50 & 58.20 & 65.80 & 38.50 & 45.60 & 27.20 \\
\midrule
BFCL v3 & Multi-Turn & 58.50 & 33.50 & 53.00 & 44.00 & \textbf{60.50} & 44.00 & 53.50 & 35.00 \\
\midrule
AceBench & Agentic & \textbf{75.00} & 32.50 & 60.00 & 63.30 & 65.00 & 40.80 & 39.10 & 63.40 \\
\midrule
\textbf{Overall} & -- & \textbf{67.28} & 61.15 & 56.04 & 58.61 & 63.07 & 47.93 & 52.26 & 50.14 \\
\bottomrule
\end{tabular}
}
\end{table}

Across our direct runs, MAVEN improves over the base GPT-OSS-120b model, with gains varying by domain. These gains are most pronounced in multi-turn and agentic settings, where structured decomposition and intermediate verification can reduce error propagation. Most external reference scores are derived from~\cite{agentscaler}, and, except AceBench, all experiments are conducted in the Function Calling (FC) evaluation setting. The observed improvements over the GPT-OSS-120b base model suggest that the MAVEN reasoning layer helps in settings where the agent must decompose a task, preserve intermediate state, and verify tool outputs before finalizing an answer. The scaffold encourages explicit validation of intermediate steps during tool invocation, which can reduce downstream error accumulation. It also provides a structured interface for selecting and sequencing tools based on the current subproblem. In the evaluated settings, this improves reliability on long-horizon tasks compared with the base model. We therefore view MAVEN as a scaffolded reasoning augmentation rather than as a new foundation model.

\subsection{Leaderboard and Representative Results}

Table~\ref{tab:leaderboard} presents representative results across a set of evaluated models on MAVEN-Bench. The results highlight the sensitivity of MAVEN-Bench to tool orchestration behavior and demonstrate the discriminative capability of the evaluation protocol.

\begin{table*}[t]
\centering
\caption{Representative model performance from direct MAVEN-Bench runs. Best results per column are highlighted in \textbf{bold}.  Pricing information corresponds to OpenRouter estimates as of October 15, 2025.}
\label{tab:leaderboard}
\footnotesize
\setlength{\tabcolsep}{5pt}
\renewcommand{\arraystretch}{1.2}
\begin{tabular}{@{}lcccccc@{}}
\toprule
\textbf{Model} & \textbf{Acc. (\%)} & \textbf{Score (/100)} & \textbf{Tool (/70)} & \textbf{Corr. (/20)} & \textbf{Appr. (/10)} & \textbf{Price} \\
\midrule
\textbf{MAVEN (GPT-OSS-120b)} & \textbf{71.0} & \textbf{95.1} & \textbf{67.4} & \textbf{18.2} & \textbf{9.5} & \$1.5 \\
Claude-Sonnet-4.5 & 70.0 & 94.2 & 67.0 & 17.7 & 9.5 & \$15 \\
Kimi-K2~\cite{kimi} & 57.0 & 92.4 & 65.5 & 17.7 & 9.3 & \$5 \\
Grok-4~\cite{xai2025grok4} & 55.0 & 88.2 & 63.0 & 16.8 & 8.4 & \$15 \\
GPT-OSS-120b~\cite{oss} & 48.0 & 88.9 & 62.6 & 17.5 & 8.7 & \$0.9 \\
GLM-4.5~\cite{glm45} & 43.0 & 88.0 & 62.3 & 16.9 & 8.7 & \$3 \\
o4-mini~\cite{o4} & 38.0 & 76.7 & 53.6 & 15.7 & 7.5 & \$9 \\
GPT-5~\cite{gpt5} & 32.0 & 61.1 & 43.3 & 12.6 & 5.2 & \$15 \\
\bottomrule
\end{tabular}
\end{table*}

\subsection{Analysis and Observed Failure Modes}

Across evaluated systems, several recurring failure modes emerge that highlight fundamental limitations in current tool-augmented reasoning architectures. A primary issue is incorrect tool selection, where models choose numerically unstable solvers or inappropriate symbolic routines, leading to degraded tool-usage performance even when the final answer remains partially recoverable. In addition, many agents fail to perform explicit intermediate verification, omitting critical checks such as sign validation or second-derivative conditions, which increases susceptibility to subtle correctness errors and reduces methodological reliability. Protocol violations are also observed, particularly among high-capacity unconstrained models that occasionally deviate from the single-call constraint; while the reconstruction pipeline partially mitigates these effects, residual penalties remain due to imperfect trace recovery. Finally, numerical instability constitutes a significant challenge, as adversarial parameter regimes in MAVEN-Bench expose conditioning issues that require both detection and adaptive handling. Collectively, these failure modes underscore the importance of structured verification, disciplined tool orchestration, and conditioning-aware reasoning.

\subsection{Paired Cross-Benchmark Robustness with MAVEN}

\begin{figure*}[t]
\centering
\begin{tikzpicture}
\begin{groupplot}[
    group style={
        group size=2 by 2,
        horizontal sep=1.5cm,
        vertical sep=1.5cm
    },
    width=0.47\textwidth,
    height=0.30\textwidth,
    xmin=5.8, xmax=15.2,
    ymin=0, ymax=105,
    xtick={6,8,10,15},
    ytick={0,25,50,75,100},
    xlabel={Steps},
    ylabel={Accuracy (\%)},
    xlabel style={font=\small},
    ylabel style={font=\small},
    tick label style={font=\scriptsize},
    title style={font=\small\bfseries, yshift=-0.5ex},
    grid=both,
    major grid style={draw=gray!25, line width=0.3pt},
    minor grid style={draw=gray!12, line width=0.2pt},
    minor tick num=1,
    legend style={
        font=\scriptsize,
        draw=none,
        fill=none,
        at={(1.07,1.18)},
        anchor=south,
        legend columns=2
    },
    every axis plot/.append style={smooth, line width=1.4pt, mark size=2.4pt}
]

\nextgroupplot[title={GPT-5}]
\addplot[
    color=red!75!black,
    mark=*,
] coordinates {
    (6,14) (7,23) (8,15) (9,7) (10,5) (15,0)
};
\addlegendentry{Without MAVEN}
\addplot[
    color=blue!75!black,
    mark=square*,
] coordinates {
    (6,100) (7,69) (8,84) (9,71) (10,36) (15,0)
};
\addlegendentry{With MAVEN}

\nextgroupplot[title={GLM-4.5}]
\addplot[
    color=red!75!black,
    mark=*,
] coordinates {
    (6,99) (7,83) (8,83) (9,38) (10,31) (15,16)
};
\addplot[
    color=blue!75!black,
    mark=square*,
] coordinates {
    (6,100) (7,84) (8,84) (9,92) (10,57) (15,33)
};

\nextgroupplot[title={Llama-4-Maverick}]
\addplot[
    color=red!75!black,
    mark=*,
] coordinates {
    (6,14) (7,0) (8,7) (9,28) (10,21) (15,0)
};
\addplot[
    color=blue!75!black,
    mark=square*,
] coordinates {
    (6,85.7) (7,84) (8,84) (9,100) (10,84) (15,50)
};

\nextgroupplot[title={Grok-4}]
\addplot[
    color=red!75!black,
    mark=*,
] coordinates {
    (6,85) (7,70) (8,84) (9,64) (10,47) (15,34)
};
\addplot[
    color=blue!75!black,
    mark=square*,
] coordinates {
    (6,100) (7,80) (8,100) (9,83) (10,54) (15,50)
};

\end{groupplot}
\end{tikzpicture}
\caption{Accuracy on MAVEN-Bench as a function of the minimum number of reasoning steps required for solution. Across the evaluated models, performance generally degrades as problem complexity increases; however, MAVEN reduces this degradation in the evaluated settings and yields stronger long-horizon robustness.}
\label{fig:grid}
\end{figure*}
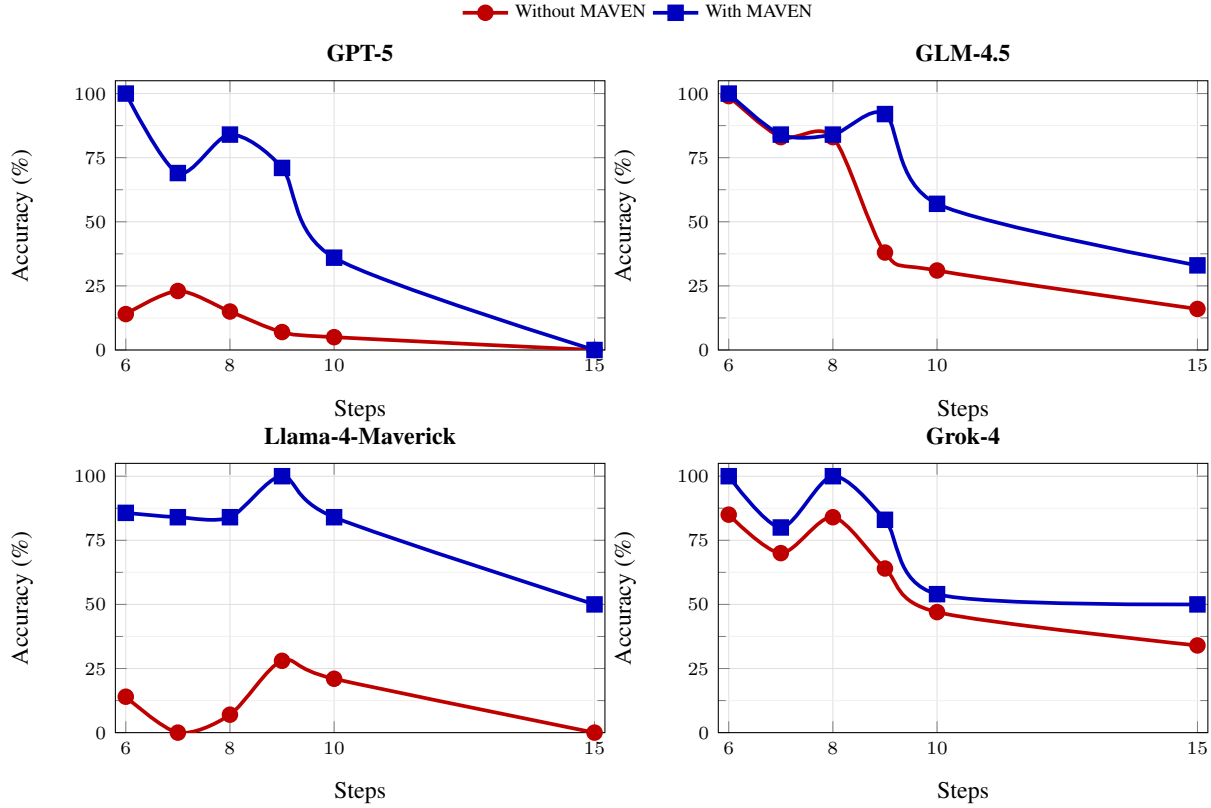

To study whether the MAVEN scaffold transfers beyond a single benchmark, we integrate the reasoning layer with several base models and evaluate performance on BFCL Multi-Turn and MAVEN-Bench (Table~\ref{tab:maven-generalization}). We describe these results as cross-benchmark robustness: the experiments test transfer across the evaluated benchmark settings. The results indicate that MAVEN improves performance over the corresponding base models in the evaluated configurations. The largest gains occur on MAVEN-Bench, where tasks require explicit decomposition, intermediate verification, and tool-state persistence. These findings support the hypothesis that structured reasoning scaffolds can reduce error accumulation in long-horizon tool-use settings. We further analyze performance as a function of problem complexity (Figure~\ref{fig:grid}), measured by the minimum number of reasoning steps required to reach a solution. Accuracy generally decreases as the number of required steps increases. Models augmented with MAVEN show a slower degradation rate, suggesting improved robustness on longer traces.

\begin{table}[t]
\centering
\caption{Available paired comparisons with and without MAVEN.}
\label{tab:maven-generalization}
\footnotesize
\setlength{\tabcolsep}{5pt}
\renewcommand{\arraystretch}{1.2}
\begin{tabularx}{\columnwidth}{@{}lXc@{}}
\toprule
\textbf{Benchmark} & \textbf{Model} & \textbf{Score} \\
\midrule
\multirow{4}{*}{\shortstack[l]{BFCL\\Multi-Turn}}
& GPT-5 & 33.5 \\
& MAVEN (GPT-5) & \textbf{51.5} \\
& Llama-4-Maverick & 23.5 \\
& MAVEN (Llama-4-Maverick) & \textbf{46} \\
\midrule
\multirow{8}{*}{MAVEN-Bench}
& GPT-5 & 32 \\
& MAVEN (GPT-5) & \textbf{66} \\
& GLM-4.5 & 43 \\
& MAVEN (GLM-4.5) & \textbf{59} \\
& Grok-4 & 55 \\
& MAVEN (Grok-4) & \textbf{69} \\
& Llama-4-Maverick & 6 \\
& MAVEN (Llama-4-Maverick) & \textbf{54} \\
\bottomrule
\end{tabularx}
\end{table}

\section{Discussion}

The results reveal a gap between benchmark performance and robust reasoning in agentic systems: models that perform well on standard benchmarks often degrade in settings requiring sustained decomposition, state management, and explicit verification, suggesting that apparent gains may reflect adaptation to fixed evaluation formats rather than generalizable reasoning ability. A key failure pattern is the mismatch between partially correct intermediate reasoning and incorrect outcomes, where small errors in tool selection, numerical stability, or missing verification steps compound over long horizons. MAVEN improves performance in the evaluated settings by externalizing reasoning into structured, composable steps with explicit verification, reducing error propagation and improving trace reliability, particularly on complex, multi-step tasks. However, these findings should be interpreted with several limitations: MAVEN-Bench focuses on mathematical and physical domains and may not capture the full diversity of real-world agentic tasks; the constrained execution protocol (e.g., single-step tool calls) may disadvantage models optimized for more flexible reasoning; evaluation partially relies on an LLM-based judge, which may introduce bias despite rubric calibration; and results are reported on a limited set of models and configurations, leaving broader generalization for future work. Together, the results suggest that process-aware evaluation is necessary to assess reliability in long-horizon settings.

\section{Conclusion}

In this work, we study agentic tool-calling systems through the lens of compositional reasoning and process-aware evaluation. We introduce \textbf{MAVEN-Bench}, a benchmark for long-horizon mathematical and physical reasoning with explicit tool use, persistent intermediate state, and verification checkpoints. MAVEN-Bench is designed to reveal failures that are not always visible from final-answer scoring alone, including incorrect tool choice, missing verification, numerical instability, and protocol violations. We also evaluate \textbf{MAVEN}, a scaffolded reasoning framework that augments LLMs with structured decomposition, explicit intermediate verification, and adaptive tool orchestration. In our direct evaluations, MAVEN improves over the corresponding GPT-OSS-120b base model and shows stronger performance on MAVEN-Bench-style multi-step reasoning tasks. Overall, the work argues for evaluating agentic systems through both outcomes and traces, where reliable tool-using agents should not only produce correct final answers, but also follow auditable, verifiable, and compositional reasoning processes.

\clearpage
\bibliography{main}
\bibliographystyle{icml2026}

\end{document}